# Kernel Methods on the Riemannian Manifold of Symmetric Positive Definite Matrices

Sadeep Jayasumana[1,2], Richard Hartley[1,2], Mathieu Salzmann[2], Hongdong Li[1], and Mehrtash Harandi[2]

[1]Australian National University, Canberra    [2]NICTA, Canberra*

sadeep.jayasumana@anu.edu.au

## Abstract

*Symmetric Positive Definite (SPD) matrices have become popular to encode image information. Accounting for the geometry of the Riemannian manifold of SPD matrices has proven key to the success of many algorithms. However, most existing methods only approximate the true shape of the manifold locally by its tangent plane. In this paper, inspired by kernel methods, we propose to map SPD matrices to a high dimensional Hilbert space where Euclidean geometry applies. To encode the geometry of the manifold in the mapping, we introduce a family of provably positive definite kernels on the Riemannian manifold of SPD matrices. These kernels are derived from the Gaussian kernel, but exploit different metrics on the manifold. This lets us extend kernel-based algorithms developed for Euclidean spaces, such as SVM and kernel PCA, to the Riemannian manifold of SPD matrices. We demonstrate the benefits of our approach on the problems of pedestrian detection, object categorization, texture analysis, 2D motion segmentation and Diffusion Tensor Imaging (DTI) segmentation.*

## 1. Introduction

Many mathematical entities in computer vision do not form vector spaces, but reside on non-linear manifolds. For instance, 3D rotation matrices form the $SO(3)$ group, linear subspaces of the Euclidean space form the Grassmann manifold, and normalized histograms form the unit $n$-sphere $S^n$. Symmetric positive definite (SPD) matrices are another class of entities lying on a Riemannian manifold. Examples of SPD matrices in computer vision include covariance region descriptors [19], diffusion tensors [13] and structure tensors [8].

Despite the abundance of such manifold-valued data, computer vision algorithms are still primarily developed for data points lying in Euclidean space ($\mathbb{R}^n$). Applying these algorithms directly to points on non-linear manifolds, and thus neglecting the geometry of the manifold, often yields poor accuracy and undesirable effects, such as the swelling of diffusion tensors in the case of SPD matrices [2, 13].

Recently, many attempts have been made to generalize algorithms developed for $\mathbb{R}^n$ to Riemannian manifolds [20, 8]. The most common approach consists in computing the tangent space to the manifold at the mean of the data points to obtain a Euclidean approximation of the manifold [20]. The logarithmic and exponential maps are then iteratively used to map points from the manifold to the tangent space, and vice-versa. Unfortunately, the resulting algorithms suffer from two drawbacks: The iterative use of the logarithmic and exponential maps makes them computationally expensive, and, more importantly, they only approximate true distances on the manifold by Euclidean distances on the tangent space.

To overcome this limitation, one could think of following the idea of kernel methods, and embed the manifold in a high dimensional Reproducing Kernel Hilbert Space (RKHS), to which many Euclidean algorithms can be generalized. In $\mathbb{R}^n$, kernel methods have proven effective for many computer vision tasks. The mapping to a RKHS relies on a kernel function, which, according to Mercer's theorem, must be positive definite. The Gaussian kernel is perhaps the most popular example of such positive definite kernels on $\mathbb{R}^n$. It would therefore seem natural to adapt this kernel to account for the geometry of Riemannian manifolds by replacing the Euclidean distance in the Gaussian kernel with the geodesic distance on the manifold. However, a kernel derived in this manner is *not* positive definite in general.

In this paper, we aim to generalize the successful and powerful kernel methods to manifold-valued data. In particular, we focus on the space of $d \times d$ SPD matrices, $Sym_d^+$, which, endowed with an appropriate metric, forms a Riemannian manifold. We present a family of provably positive definite kernels on $Sym_d^+$ derived by accounting for the non-linear geometry of the manifold.

More specifically, we propose a theoretical framework to analyze the positive definiteness of the Gaussian kernel generated by a distance function on any non-linear manifold. Using this framework, we show that a family of metrics on $Sym_d^+$ define valid positive definite Gaussian kernels when replacing the Euclidean distance with the dis-

*NICTA is funded by the Australian Government as represented by the Department of Broadband, Communications and the Digital Economy and the ARC through the ICT Centre of Excellence program.

This work was supported in part by an ARC grant.

tance corresponding to these metrics. A notable special case of such metrics is the log-Euclidean metric, which has been shown to define a true geodesic distance on $Sym_d^+$. We demonstrate the benefits of our manifold-based kernel by exploiting it in four different algorithms. Our experiments show that the resulting manifold kernel methods outperform the corresponding Euclidean kernel methods, as well as the manifold methods that use tangent space approximations.

## 2. Related Work

SPD matrices find a variety of applications in computer vision. For instance, covariance region descriptors are used in object detection [20], texture classification [19], object tracking, action recognition and face recognition [9]. Diffusion Tensor Imaging (DTI) was one of the pioneering fields for the development of non-linear algorithms on $Sym_d^+$ [13, 2]. In optical flow estimation and motion segmentation, structure tensors are often employed to encode important image features, such as texture and motion [8].

In recent years, several optimization algorithms on manifolds have been proposed for $Sym_d^+$. In particular, LogitBoost on a manifold was introduced for binary classification [20]. This algorithm has the drawbacks of approximating the manifold by tangent spaces and not scaling with the number of training samples due to the iterative use of exponential and logarithmic maps. Making use of our positive definite kernels yields more efficient and accurate classification algorithms on non-linear manifolds. Dimensionality reduction and clustering on $Sym_d^+$ was demonstrated in [8] with Riemannian versions of the Laplacian Eigenmaps (LE), Locally Linear Embedding (LLE) and Hessian LLE (HLLE). Clustering was performed in a low dimensional space after dimensionality reduction, which does not necessarily preserve all the information in the original data distribution. We instead utilize our kernels to perform clustering in a higher dimensional RKHS that embeds $Sym_d^+$.

The use of kernels on $Sym_d^+$ has previously been advocated for locality preserving projections [10] and sparse coding [9]. In the first case, the kernel, derived from the affine-invariant distance, is not positive definite in general [10]. In the second case, the kernel uses the Stein divergence, which is not a true geodesic distance, as the distance measure and is positive definite only for some values of the Gaussian bandwidth parameter $\sigma$ [9]. For all kernel methods, the optimal choice of $\sigma$ largely depends on the data distribution and hence constraints on $\sigma$ are not desirable. Moreover, many popular automatic model selection methods require $\sigma$ to be continuously variable [5].

Other than for satisfying Mercer's theorem to generate a valid RKHS, positive definiteness of the kernel is a required condition for the convergence of many kernel based algorithms. For instance, the Support Vector Machine (SVM) learning problem is convex only when the kernel is positive definite [14]. Similarly, positive definiteness of all participating kernels is required to guarantee the convexity in Multiple Kernel Learning (MKL) [22]. Although theories have been proposed to exploit non-positive definite kernels [12, 23], they have not experienced a widespread success. Many of these methods first enforce positive definiteness of the kernel by flipping or shifting its negative eigenvalues [23]. As a consequence, they result in a loss of information and become inapplicable with large sized kernels that are not uncommon in learning problems.

Recently, mean-shift clustering with a positive definite heat kernel on Riemannian manifolds was introduced [4]. However, due to the mathematical complexity of the kernel function, computing it is not tractable and hence only an approximation of the true kernel was used in the algorithm.

Here, we introduce a family of provably positive definite kernels on $Sym_d^+$, and show their benefits in various kernel-based algorithms and on several computer vision tasks.

## 3. Background

In this section, we introduce some notions of Riemannian geometry on the manifold of SPD matrices, and discuss the use of kernel methods on non-linear manifolds.

### 3.1. The Riemannian Manifold of SPD Matrices

A differentiable manifold $\mathcal{M}$ is a topological space that is locally similar to Euclidean space and has a globally defined differential structure. The tangent space at a point $p$ on the manifold, $T_p\mathcal{M}$, is a vector space that consists of the tangent vectors of all possible curves passing through $p$.

A Riemannian manifold is a differentiable manifold equipped with a smoothly varying inner product on each tangent space. The family of inner products on all tangent spaces is known as the *Riemannian metric* of the manifold. It enables to define various geometric notions on the manifold such as the angle between two curves, or the length of a curve. The *geodesic distance* between two points on the manifold is defined as the length of the shortest curve connecting the two points. Such shortest curves are known as *geodesics* and are analogous to straight lines in $\mathbb{R}^n$.

The space of $d \times d$ SPD matrices, $Sym_d^+$, is mostly studied when endowed with a Riemannian metric and thus forms a Riemannian manifold [13, 1]. In such a case, the geodesic distance induced by the Riemannian metric is a more natural measure of dissimilarity between two SPD matrices than the Euclidean distance. Although a number of metrics[1] on $Sym_d^+$ have been recently proposed to capture its non-linearity, not all of them arise from a smoothly varying inner product on tangent spaces and thus define a true geodesic distance. The two most widely used distance measures are the affine-invariant distance [13] and the log-Euclidean distance [2]. The main reason for their popularity is that they are true geodesic distances induced by Riemannian metrics. For a review of metrics on $Sym_d^+$, the reader is referred to [7].

---
[1]The term *metric* refers to a distance function that satisfies the four metric axioms, while *Riemannian metric* refers to a family of inner products.

### 3.2. Kernel Methods on Non-linear Manifolds

Kernel methods in $\mathbb{R}^n$ have proven extremely effective in machine learning and computer vision to explore non-linear patterns in data. The fundamental idea of kernel methods is to map the input data to a high (possibly infinite) dimensional feature space to obtain a richer representation of the data distribution.

This concept can be generalized to non-linear manifolds as follows: Each point $x$ on a non-linear manifold $\mathcal{M}$ is mapped to a feature vector $\phi(x)$ in a Hilbert space $\mathcal{H}$, the Cauchy completion of the space spanned by real-valued functions defined on $\mathcal{M}$. A kernel function $k : (\mathcal{M} \times \mathcal{M}) \to \mathbb{R}$ is used to define the inner product on $\mathcal{H}$, thus making it a Reproducing Kernel Hilbert Space (RKHS). According to Mercer's theorem, however, only positive definite kernels define valid RKHS.

Since, in general, Riemannian manifolds are non-linear, many algorithms designed for $\mathbb{R}^n$ cannot directly be utilized on them. To overcome this, most existing methods map the points on the manifold to the tangent space at one point (usually the mean point), thus obtaining a Euclidean representation of the manifold-valued data. Unfortunately, such a mapping does not globally preserve distances and hence yields a poor representation of the original data distribution. In contrast, many algorithms on $\mathbb{R}^n$ can be directly generalized to Hilbert spaces, where vector norms and inner products are defined. As a consequence, there are two advantages in using kernel functions to embed a manifold in an RKHS. First, the mapping transforms the non-linear manifold into a (linear) Hilbert space, thus making it possible to utilize algorithms designed for $\mathbb{R}^n$ with manifold-valued data. Second, as evidenced by the theory of kernel methods on $\mathbb{R}^n$, it yields a much richer representation of the original data distribution. These benefits, however, depend on the condition that the kernel be positive definite. We address this in the next section.

## 4. Positive Definite Kernels on Manifolds

In this section, we first present a general theory to analyze the positive definiteness of Gaussian kernels defined on manifolds and then introduce a family of provably positive definite kernels on $Sym_d^+$.

### 4.1. The Gaussian Kernel on a Metric Space

The Gaussian radial basis function (RBF) has proven very effective in Euclidean space as a positive definite kernel for kernel based algorithms. It maps the data points to an infinite dimensional Hilbert space, which, intuitively, yields a very rich representation. In $\mathbb{R}^n$, the Gaussian kernel can be expressed as $k_G(\mathbf{x}_i, \mathbf{x}_j) := \exp(\|\mathbf{x}_i - \mathbf{x}_j\|^2/2\sigma^2)$, which makes use of the Euclidean distance between two data points $\mathbf{x}_i$ and $\mathbf{x}_j$. To define a kernel on a Riemannian manifold, we would like to replace the Euclidean distance by a more accurate geodesic distance on the manifold. However, not all geodesic distances yield positive definite kernels.

We now state our main theorem, which states sufficient and necessary conditions to obtain a positive definite Gaussian kernel from a distance function.

**Theorem 4.1.** *Let $(M, d)$ be a metric space and define $k : (M \times M) \to \mathbb{R}$ by $k(x_i, x_j) := \exp(-d^2(x_i, x_j)/2\sigma^2)$. Then, $k$ is a positive definite kernel for all $\sigma > 0$ if and only if there exists an inner product space $\mathcal{V}$ and a function $\psi : M \to \mathcal{V}$ such that, $d(x_i, x_j) = \|\psi(x_i) - \psi(x_j)\|_\mathcal{V}$.*

*Proof.* The proof of Theorem 4.1 follows a number of steps detailed below. We start with the definition of positive and negative definite functions [3].

**Definition 4.2.** *Let $\mathcal{X}$ be a nonempty set. A function $f : (\mathcal{X} \times \mathcal{X}) \to \mathbb{R}$ is called a **positive (resp. negative) definite kernel** if and only if $f$ is symmetric and*

$$\sum_{i,j=1}^m c_i c_j f(x_i, x_j) \geq 0 \quad (resp. \leq 0)$$

*for all $m \in \mathbb{N}, \{x_1, \ldots, x_m\} \subseteq \mathcal{X}$ and $\{c_1, ..., c_m\} \subseteq \mathbb{R}$, with $\sum_{i=1}^m c_i = 0$ in the negative definite case.*

Given this definition, we make use of the following important theorem due mainly to Schoenberg [16].

**Theorem 4.3.** *Let $\mathcal{X}$ be a nonempty set and $f : (\mathcal{X} \times \mathcal{X}) \to \mathbb{R}$ be a function. The kernel $\exp(-tf(x_i, x_j))$ is positive definite for all $t > 0$ if and only if $f$ is negative definite.*

*Proof.* We refer the reader to Chapter 3, Theorem 2.2 of [3] for a detailed proof of this theorem. □

Although the origin of this theorem dates back to 1938 [16], it has received little attention in the computer vision community. Theorem 4.3 implies that positive definiteness of the Gaussian kernel induced by a distance is equivalent to negative definiteness of the squared distance function. Therefore, to prove the positive definiteness of $k$ in Theorem 4.1, we only need to prove that $d^2$ is negative definite. We formalize this in the next theorem:

**Theorem 4.4.** *Let $\mathcal{X}$ be a nonempty set, $\mathcal{V}$ be an inner product space, and $\psi : \mathcal{X} \to \mathcal{V}$ be a function. Then, $f : (\mathcal{X} \times \mathcal{X}) \to \mathbb{R}$ defined by $f(x_i, x_j) := \|\psi(x_i) - \psi(x_j)\|_\mathcal{V}^2$, is negative definite.*

| Metric Name | Formula | Geodesic Distance | Positive Definite Gaussian Kernel $\forall \sigma > 0$ |
|---|---|---|---|
| **Log-Euclidean** | $\|\log(\mathbf{S}_1) - \log(\mathbf{S}_2)\|_F$ | **Yes** | **Yes** |
| Affine-Invariant | $\|\log(\mathbf{S}_1^{-1/2}\mathbf{S}_2\mathbf{S}_1^{-1/2})\|_F$ | Yes | No |
| Cholesky | $\|\text{chol}(\mathbf{S}_1) - \text{chol}(\mathbf{S}_2)\|_F$ | No | Yes |
| Power-Euclidean | $\frac{1}{\alpha}\|\mathbf{S}_1^\alpha - \mathbf{S}_2^\alpha\|_F$ | No | Yes |
| Root Stein Divergence | $\left[\log\det\left(\frac{1}{2}\mathbf{S}_1 + \frac{1}{2}\mathbf{S}_2\right) - \frac{1}{2}\log\det(\mathbf{S}_1\mathbf{S}_2)\right]^{1/2}$ | No | No |

Table 1: **Properties of different metrics on** $Sym_d^+$. We analyze positive definiteness of Gaussian kernels generated by different metrics. While Theorem 4.1 applies to the metrics claimed to generate positive definite Gaussian kernels, examples of non-positive definite Gaussian kernels exist for other metrics.

*Proof.* Based on Definition 4.2, we need to prove that $\sum_{i,j=1}^m c_i c_j f(x_i, x_j) \leq 0$ for all $m \in \mathbb{N}$, $\{x_1, \ldots, x_m\} \subseteq \mathcal{X}$ and $\{c_1, \ldots, c_m\} \subseteq \mathbb{R}$ with $\sum_{i=1}^m c_i = 0$.

$$\sum_{i,j=1}^m c_i c_j f(x_i, x_j) = \sum_{i,j=1}^m c_i c_j \|\psi(x_i) - \psi(x_j)\|_\mathcal{V}^2$$

$$= \sum_{i,j=1}^m c_i c_j \langle \psi(x_i) - \psi(x_j), \psi(x_i) - \psi(x_j)\rangle_\mathcal{V}$$

$$= \sum_{j=1}^m c_j \sum_{i=1}^m c_i \langle \psi(x_i), \psi(x_i)\rangle_\mathcal{V}$$
$$- 2 \sum_{i,j=1}^m c_i c_j \langle \psi(x_i), \psi(x_j)\rangle_\mathcal{V}$$
$$+ \sum_{i=1}^m c_i \sum_{j=1}^m c_j \langle \psi(x_j), \psi(x_j)\rangle_\mathcal{V}$$

$$= -2 \sum_{i,j=1}^m c_i c_j \langle \psi(x_i), \psi(x_j)\rangle_\mathcal{V}$$

$$= -2 \left\|\sum_{i=1}^m c_i \psi(x_i)\right\|_\mathcal{V}^2 \leq 0. \qquad \square$$

Combining Theorem 4.4 and Theorem 4.3 proves the forward direction of Theorem 4.1, which is useful for the work presented in this paper. The converse can be proved by combining Theorem 4.3 and Proposition 3.2 in Chapter 3 of [3], we omit the details due to space limitations. $\square$

### 4.2. Kernels on $Sym_d^+$

We now discuss the different metrics on $Sym_d^+$ that can be used to define positive definite Gaussian kernels. In particular, we focus on the log-Euclidean distance which is a true geodesic distance on $Sym_d^+$ [2].

The log-Euclidean distance for $Sym_d^+$ was derived by exploiting the Lie group structure of $Sym_d^+$ under the group operation $\mathbf{X}_i \odot \mathbf{X}_j := \exp(\log(\mathbf{X}_i) + \log(\mathbf{X}_j))$ for $\mathbf{X}_i, \mathbf{X}_j \in Sym_d^+$ where $\exp(\cdot)$ and $\log(\cdot)$ denote the usual matrix exponential and logarithm operators (not to be confused with exponential and logarithmic maps of the log-Euclidean Riemannian metric, which are point dependent and take more complex forms [1]). Under the log-Euclidean framework, a geodesic connecting $\mathbf{X}_i, \mathbf{X}_j \in Sym_d^+$ is defined as $\gamma(t) = \exp((1-t)\log(\mathbf{X}_i) + t\log(\mathbf{X}_j))$ for $t \in [0, 1]$. The geodesic distance between $\mathbf{X}_i$ and $\mathbf{X}_j$ can be expressed as

$$d_g(\mathbf{X}_i, \mathbf{X}_j) = \|\log(\mathbf{X}_i) - \log(\mathbf{X}_j)\|_F , \qquad (1)$$

where $\|\cdot\|_F$ denotes the Frobenius matrix norm induced by the Frobenius matrix inner product $\langle \cdot, \cdot\rangle_F$.

The main reason to exploit the log-Euclidean distance in our experiments is that it defines a true geodesic distance that has proven an effective distance measure on $Sym_d^+$. Furthermore, it yields a positive definite Gaussian kernel as stated in the following corollary to Theorem 4.1:

**Corollary 4.5** (Theorem 4.1). *Let* $k_R : (Sym_d^+ \times Sym_d^+) \to \mathbb{R} : k_R(\mathbf{X}_i, \mathbf{X}_j) := \exp(-d_g^2(\mathbf{X}_i, \mathbf{X}_j)/2\sigma^2)$, *with* $d_g(\mathbf{X}_i, \mathbf{X}_j) = \|\log(\mathbf{X}_i) - \log(\mathbf{X}_j)\|_F$. *Then,* $k_R$ *is a positive definite kernel for all* $\sigma \in \mathbb{R}$.

*Proof.* Directly follows Theorem 4.1 with the Frobenius matrix inner product. $\square$

A number of other metrics have been proposed for $Sym_d^+$ [7]. The definitions and properties of these metrics are summarized in Table 1. Note that only some of them were derived by considering the Riemannian geometry of the manifold and hence define true geodesic distances. Similar to the log-Euclidean metric, from Theorem 4.1, it directly follows that the Cholesky and power-Euclidean metrics also define positive definite Gaussian kernels for all values of $\sigma$. Note that some metrics may yield a positive definite Gaussian kernel for some value of $\sigma$ only. This, for instance, was shown in [18] for the root Stein divergence metric. No such result is known for the affine-invariant metric. Constraints on $\sigma$ are nonetheless undesirable, since $\sigma$ should reflect the data distribution and automatic model selection algorithms require $\sigma$ to be continuously variable [5].

# 5. Kernel-based Algorithms on $Sym_d^+$

A major advantage of being able to compute positive definite kernels on a Riemannian manifold is that it directly allows us to make use of algorithms developed for $\mathbb{R}^n$, while still accounting for the geometry of the manifold. In this section, we discuss the use of four kernel-based algorithms on $Sym_d^+$. The resulting algorithms can be thought of as generalizations of the original ones to non-linear manifolds. In the following, we use $k(.,.)$, $\mathcal{H}$ and $\phi(\mathbf{X})$ to denote the kernel function defined in Theorem 4.1, the RKHS generated by $k$, and the feature vector in $\mathcal{H}$ to which $\mathbf{X} \in Sym_d^+$ is mapped, respectively. Although we use $\phi(\mathbf{X})$ for explanation purposes, following the *kernel trick*, it never needs be explicitly computed.

## 5.1. Kernel Support Vector Machines on $Sym_d^+$

We first consider the case of using kernel SVM for binary classification on a manifold. Given a set of training examples $\{(\mathbf{X}_i, y_i)\}_1^m$, where $\mathbf{X}_i \in Sym_d^+$ and the label $y_i \in \{-1, 1\}$, kernel SVM searches for a hyperplane in $\mathcal{H}$ that separates the feature vectors of the positive and negative classes with maximum margin. The class of a test point $\mathbf{X}$ is determined by the position of the feature vector $\phi(\mathbf{X})$ in $\mathcal{H}$ relative to the separating hyperplane. Classification with kernel SVM can be done very fast, since it only requires to evaluate the kernel at the *support vectors*.

Kernel SVM on $Sym_d^+$ is much simpler to implement and less computationally demanding in both training and testing phases than the current state-of-the-art binary classification algorithms on $Sym_d^+$, such as LogitBoost on a manifold [20], which involves iteratively combining weak learners on different tangent spaces. Weighted mean calculation in LogitBoost on a manifold involves an extremely expensive gradient descent procedure at each boosting iteration, which makes the algorithm scale poorly with the number of training samples. Furthermore, while LogitBoost learns classifiers on tangent spaces used as Euclidean approximates of the manifold, our approach makes use of a rich high dimensional feature space. As will be shown in our experiments, this yields better classification results.

## 5.2. Multiple Kernel Learning on $Sym_d^+$

The core idea of Multiple Kernel Learning (MKL) is to combine kernels computed from different descriptors (e.g., image features) to obtain a kernel that optimally separates two classes for a given classifier. Here, we follow the formulation of [22], and make use of an SVM classifier. As a feature selection method, MKL has proven more effective than conventional feature selection methods such as wrappers, filters and boosting [21].

More specifically, given training examples $\{(x_i, y_i)\}_1^m$, where $x_i \in \mathcal{X}$, $y_i \in \{-1, 1\}$, and a set of descriptor generating functions $\{g_j\}_1^N$ where $g_j : \mathcal{X} \to Sym_d^+$, we seek to learn a binary classifier $f : \mathcal{X} \to \{-1, 1\}$ by selecting and optimally combining the different descriptors generated by $g_1, \ldots, g_N$. Let $\mathbf{K}^{(j)}$ be the kernel matrix generated by $g_j$ and $k$ as $\mathbf{K}_{pq}^{(j)} = k(g_j(x_p), g_j(x_q))$. The combined kernel can be expressed as $\mathbf{K}^* = \sum_j \lambda_j \mathbf{K}^{(j)}$, where $\lambda_j \geq 0$ for $j = 1 \ldots N$ guarantees the positive definiteness of $\mathbf{K}^*$. The weights $\boldsymbol{\lambda}$ can be learned using a min-max optimization procedure with an $L_1$ regularizer on $\boldsymbol{\lambda}$ to obtain a sparse combination of kernels. For more details, we refer the reader to [22] and [21]. Note that convergence of MKL is only guaranteed if all the kernels are positive definite.

## 5.3. Kernel PCA on $Sym_d^+$

We now describe the key concepts of kernel PCA on $Sym_d^+$. Kernel PCA is a non-linear dimensionality reduction method [17]. Since it works in feature space, kernel PCA may, however, extract a number of dimensions that exceeds the dimensionality of the input space. Kernel PCA proceeds as follows: All points $\mathbf{X}_i \in Sym_d^+$ of a given dataset $\{\mathbf{X}_i\}_{i=1}^m$ are mapped to feature vectors in $\mathcal{H}$, thus yielding the transformed set, $\{\phi(\mathbf{X}_i)\}_{i=1}^m$. The covariance matrix of this transformed set is then computed, which really amounts to computing the kernel matrix of the original data using the function $k$. An $l$-dimensional representation of the data is obtained by computing the eigenvectors of the kernel matrix. This representation can be thought of as a Euclidean representation of the original manifold-valued data. However, owing to our kernel, it was obtained by accounting for the geometry of $Sym_d^+$.

## 5.4. Kernel $k$-means on $Sym_d^+$

For clustering problems, we propose to make use of kernel $k$-means on $Sym_d^+$. Kernel $k$-means maps points to a high-dimensional Hilbert space and performs $k$-means on the resulting feature space [17]. More specifically, a given dataset $\{\mathbf{X}_i\}_{i=1}^m$, with each $\mathbf{X}_i \in Sym_d^+$, is clustered into a pre-defined number of groups in $\mathcal{H}$, such that the sum of the squared distances from each $\phi(\mathbf{X}_i)$ to the nearest cluster center is minimum. The resulting clusters can then act as classes for the $\{\mathbf{X}_i\}_{i=1}^m$.

The unsupervised clustering method on $Sym_d^+$ proposed in [8] clusters points in a low dimensional space after dimensionality reduction on the manifold. In contrast, our method performs clustering in a high dimensional RKHS which, intuitively, better represents the data distribution.

# 6. Applications and Experiments

We now present our experimental evaluation of the kernel methods on $Sym_d^+$ described in Section 5. In the remainder of this section, we use *Riemannian kernel* and *Euclidean kernel* to refer to the kernel defined in Corollary 4.5 and the standard Euclidean Gaussian kernel, respectively.

## 6.1. Pedestrian Detection

We first demonstrate the use of our Riemannian kernel for the task of pedestrian detection with kernel SVM and MKL on $Sym_d^+$. Let $\{(\mathbf{W}_i, y_i)\}_{i=1}^m$ be the training set, where each $\mathbf{W}_i \in \mathbb{R}^{h \times w}$ is an image window and $y_i \in \{-1, 1\}$ is the class label (background or person) of $\mathbf{W}_i$. Following [20], we use covariance descriptors computed from the feature vector $\left[x, y, |I_x|, |I_y|, \sqrt{I_x^2 + I_y^2}, |I_{xx}|, |I_{yy}|, \arctan\left(\frac{|I_x|}{|I_y|}\right)\right]$, where $x, y$ are pixel locations and $I_x, I_y, \ldots$ are intensity derivatives. The covariance matrix for an image patch of arbitrary size therefore is an $8 \times 8$ SPD matrix. In a $h \times w$ window $\mathbf{W}$, a large number of covariance descriptors can be computed from subwindows with different sizes and positions sampled from $\mathbf{W}$. We consider $N$ subwindows $\{w_j\}_{j=1}^N$ of size ranging from $h/5 \times w/5$ to $h \times w$, positioned at all possible locations. The covariance descriptor of each subwindow is normalized using the covariance descriptor of the full window to improve robustness against illumination changes. Such covariance descriptors can be computed efficiently using integral images [20].

Let $\mathbf{X}_i^{(j)} \in Sym_8^+$ denote the covariance descriptor of the $j^{\text{th}}$ subwindow of $\mathbf{W}_i$. To reduce this large number of descriptors, we pick the best 100 subwindows that do not mutually overlap by more than 75%, by ranking them according to their variance across all training samples. Since the descriptors lie on a Riemannian manifold, for each descriptor $\mathbf{X}^{(j)}$ we compute the variance across all positive training samples as

$$\text{var}(\mathbf{X}^{(j)}) = \frac{1}{m_+} \sum_{i:y_i=1} d_g^p(\mathbf{X}_i^{(j)}, \bar{\mathbf{X}}^{(j)}) \qquad (2)$$

where $m_+$ is the number of positive training samples and $\bar{\mathbf{X}}$ is the Karcher mean of $\{\mathbf{X}_i\}_{i:y_i=1}$ given by $\bar{\mathbf{X}} = \exp\left(\frac{1}{m_+} \sum_{i:y_i=1} \log(\mathbf{X}_i)\right)$ under the log-Euclidean metric. We set $p = 1$ in Eq.(2) to make the statistic less sensitive to outliers. We then use the SVM-MKL framework described in Section 5.2 to learn the final classifier, where each kernel is defined on one of the 100 selected subwindows. At test time, detection is achieved in a sliding window manner followed by a non-maxima suppression step.

To evaluate our approach, we made use of the INRIA person dataset [6]. Its training set consists of 2,416 positive windows and 1,280 person-free negative images, and its test set of 1,237 positive windows and 453 negative images. Negative windows are generated by sampling negative images [6]. We first used all positive samples and 12,800 negative samples (10 random windows from each negative image) to train an initial classifier. We used this classifier to find hard negative examples in the training images, and retrained the classifier by adding these hard examples to the training set. Cross validation was used to determine the hyperparameters including the parameter $\sigma$ of the kernel. We used the evaluation methodology of [6].

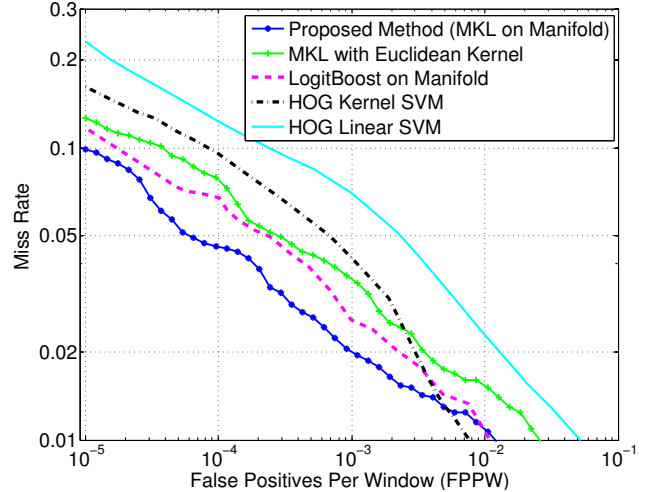

Figure 1: **Pedestrian detection.** Detection-Error tradeoff curves for the proposed manifold MKL approach and state-of-the-art methods on the INRIA dataset. The curves for the baselines were reproduced from [20].

In Figure 1, we compare the detection-error tradeoff (DET) curves of our approach and state-of-the-art methods. The curve for our method was generated by continuously varying the decision threshold of the final MKL classifier. We also evaluated our MKL framework with a Euclidean kernel. Note that the proposed MKL method with a Riemannian kernel outperforms MKL with a Euclidean kernel, as well as LogitBoost on the manifold. This suggests the importance of accounting for the geometry of the manifold.

## 6.2. Visual Object Categorization

We next tackle the problem of unsupervised object categorization. To this end, we used the ETH-80 dataset [11] which contains 8 categories with 10 objects each and 41 images per object. We used 21 randomly chosen images from each object to compute the parameter $\sigma$ and the rest to evaluate clustering accuracy. For each image, we used a single $5 \times 5$ covariance descriptor calculated from the features $[x, y, I, |I_x|, |I_y|]$, where $x, y$ are pixel locations and $I, I_x, I_y$ are intensity and derivatives. To obtain object categories, the kernel $k$-means algorithm on $Sym_5^+$ described in Section 5.4 was employed to perform clustering.

One drawback of $k$-means and its kernel counterpart is their sensitivity to initialization. To overcome this, we ran the algorithm 20 times with different random initializations and picked the iteration that converged to the minimum sum of point-to-centroid squared distances. For kernel $k$-means on $Sym_5^+$, distances in the RKHS were used. Note that we assumed $k$ to be known.

To set a benchmark, we evaluated the performance of both $k$-means and kernel $k$-means on $Sym_5^+$ with different metrics that generate positive definite Gaussian kernels (see Table 1). For the power-Euclidean metric, we used $\alpha = 0.5$, which achieved the best results in [7]. For all non-Euclidean metrics with (non-kernel) $k$-means, the Karcher mean [7] was used to compute the centroid. The results of the different methods are summarized in Table 2. Manifold kernel $k$-means with the log-Euclidean metric performs significantly better than all other methods in all test cases. These results also outperform the results with the heat kernel reported in [4]. Note, however, that [4] only considered 3 and 4 classes without mentioning which classes were used.

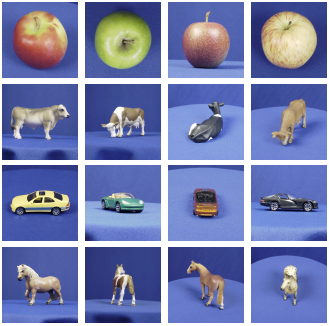

| Nb. of classes | Euclidean | | Cholesky | | Power-Euclidean | | Log-Euclidean | |
|---|---|---|---|---|---|---|---|---|
| | KM | KKM | KM | KKM | KM | KKM | KM | KKM |
| 3 | 72.50 | 79.00 | 73.17 | 82.67 | 71.33 | 84.33 | 75.00 | **94.83** |
| 4 | 64.88 | 73.75 | 69.50 | 84.62 | 69.50 | 83.50 | 73.00 | **87.50** |
| 5 | 54.80 | 70.30 | 70.80 | 82.40 | 70.20 | 82.40 | 74.60 | **85.90** |
| 6 | 50.42 | 69.00 | 59.83 | 73.58 | 59.42 | 73.17 | 66.50 | **74.50** |
| 7 | 42.57 | 68.86 | 50.36 | 69.79 | 50.14 | 69.71 | 59.64 | **73.14** |
| 8 | 40.19 | 68.00 | 53.81 | 69.44 | 54.62 | 68.44 | 58.31 | **71.44** |

Table 2: **Object categorization.** Sample images and percentages of correct clustering on the ETH-80 dataset using $k$-means (KM) and kernel $k$-means (KKM) with different metrics.

### 6.3. Texture Recognition

We then utilized our Riemannian kernel to demonstrate the effectiveness of manifold kernel PCA on texture recognition. To this end, we used the Brodatz dataset [15], which consists of 111 different $640 \times 640$ texture images. Each image was divided into four subimages of equal size, two of which were used for training and the other two for testing.

For each training image, covariance descriptors of randomly chosen 50 $128 \times 128$ windows were computed from the feature vector $[I, |I_x|, |I_y|, |I_{xx}|, |I_{yy}|]$ [19]. Kernel PCA on $Sym_5^+$ with our Riemannian kernel was then used to extract the top $l$ principal directions in the RKHS, and project the training data along those directions. Given a test image, we computed 100 covariance descriptors from random windows and projected them to the $l$ principal directions obtained during training. Each such projection was classified using a majority vote over its 5 nearest-neighbors. The class of the test image was then decided by majority voting among the 100 descriptors. Cross validation on the training set was used to determine $\sigma$. For comparison purposes, we repeated the same procedure with the Euclidean kernel. Results obtained for these kernels and different values of $l$ are presented in Table 3. The better recognition accuracy indicates that kernel PCA with the Riemannian kernel more effectively captures the information of the manifold-valued descriptors than the Euclidean kernel.

| Kernel | Classification Accuracy | | | |
|---|---|---|---|---|
| | $l=10$ | $l=11$ | $l=12$ | $l=15$ |
| Riemannian | **95.50** | **95.95** | **96.40** | **96.40** |
| Euclidean | 89.64 | 90.09 | 90.99 | 91.89 |

Table 3: **Texture recognition.** Recognition accuracies on the Brodatz dataset with $k$-NN in a $l$-dimensional Euclidean space obtained by kernel PCA.

### 6.4. Segmentation

Finally, we illustrate the use of our kernel to segment different types of images. First, we consider DTI segmentation, which is a key application area of algorithms on $Sym_d^+$. We utilized kernel $k$-means on $Sym_3^+$ with our Riemannian kernel to segment a real DTI image of the human brain. Each pixel of the input DTI image is a $3 \times 3$ SPD matrix, which can thus directly be used as input to the algorithm. The $k$ clusters obtained by the algorithm act as classes, thus yielding a segmentation of the image.

Figure 2 depicts the resulting segmentation along with the ellipsoid and fractional anisotropy representations of the original DTI image. We also show the results obtained by replacing the Riemannian kernel with the Euclidean one. Note that, up to some noise due to the lack of spatial smoothing, Riemannian kernel $k$-means was able to correctly segment the corpus callosum from the rest of the image.

We then followed the same approach to perform 2D motion segmentation. To this end, we used a spatio-temporal structure tensor directly computed on image intensities (i.e., without extracting features such as optical flow). The spatio-temporal structure tensor for each pixel is computed as $\mathbf{T} = K * (\nabla I \nabla I^T)$, where $\nabla I = (I_x, I_y, I_t)$ and $K*$ indicates convolution with the regular Gaussian kernel for smoothing. Each pixel is thus represented as a $3 \times 3$ SPD matrix and segmentation can be performed by clustering these matrices using kernel $k$-means on $Sym_3^+$.

We applied this strategy to two images taken from the Hamburg Taxi sequence. Figure 3 compares the results of kernel $k$-means with our Riemannian kernel with the results of [8] obtained by first performing LLE, LE, or HLLE on $Sym_3^+$ and then clustering in the low dimensional space.

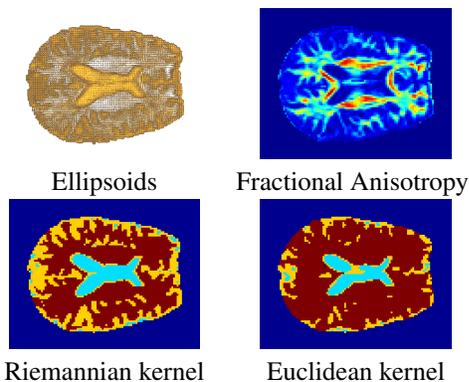

Ellipsoids    Fractional Anisotropy

Riemannian kernel    Euclidean kernel

Figure 2: **DTI segmentation.** Segmentation of the corpus callosum with kernel $k$-means on $Sym_3^+$.

Note that our approach yields a much cleaner segmentation than the baselines. This might be attributed to the fact that we perform clustering in a high dimensional feature space, whereas the baselines work in a reduced dimensional space.

## 7. Conclusion

In this paper, we have introduced a family of provably positive definite kernels on the Riemannian manifold of SPD matrices. We have shown that such kernels could be used to design Riemannian extensions of existing kernel-based algorithms, such as SVM and kernel $k$-means. Our experiments have demonstrated the benefits of these kernels over the Euclidean Gaussian kernel, as well as over other manifold-based algorithms on several tasks. Although developed for the Riemannian manifold of SPD matrices, the theory of this paper could apply to other non-linear manifolds, provided that their metrics define negative definite squared distances. We therefore intend to study which manifolds fall into this category. We also plan to investigate the positive definiteness of non-Gaussian kernels.

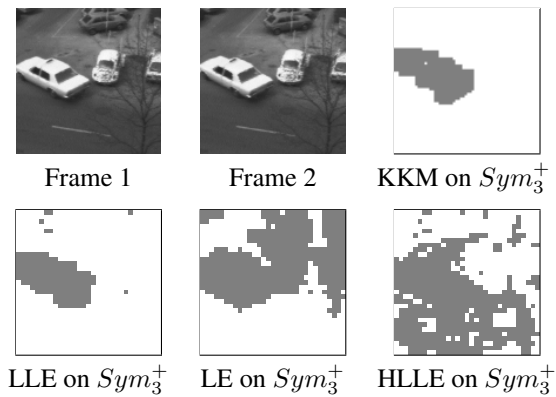

Frame 1    Frame 2    KKM on $Sym_3^+$

LLE on $Sym_3^+$    LE on $Sym_3^+$    HLLE on $Sym_3^+$

Figure 3: **2D motion segmentation.** Comparison of the segmentations obtained with kernel $k$-means with our Riemannian kernel (KKM), LLE, LE and HLLE on $Sym_3^+$. The baseline results were reproduced from [8].